\documentclass[conference,a4paper]{IEEEtran}
\IEEEoverridecommandlockouts

\usepackage[hidelinks]{hyperref}
\usepackage{xcolor}
\usepackage[table]{xcolor}
\usepackage[cmex10]{amsmath}%American Math Society(AMS) math formatting
\usepackage{amssymb,amsfonts}%AMS extra symbols and fonts
\usepackage{dblfloatfix}%fix double column figure ordering and placement

\usepackage[ruled,vlined]{algorithm2e}
\usepackage{graphicx}
\graphicspath{{Figures/PDF/}{Figures/PNG/}}

\usepackage{booktabs}
\usepackage{siunitx}
\usepackage[numbers,compress]{natbib}
\usepackage{texnames}
\usepackage{bm,bbm}
\usepackage{orcidlink}
\usepackage{array}
\usepackage[most]{tcolorbox}
\usepackage{float}

\usepackage{multirow}
\usepackage{graphicx}

\definecolor{venuegray}{gray}{0.55}
\definecolor{groupgray}{gray}{0.93}
\definecolor{lightblue}{RGB}{173, 216, 230}

\definecolor{cloudblue}{RGB}{0,170,255}
\definecolor{cloudgrey}{RGB}{150,150,150}
\definecolor{shadowpurple}{RGB}{190,100,255}
\definecolor{lightred}{RGB}{230, 145, 145}
\definecolor{lightgreen}{RGB}{125, 195, 150}

\IEEEaftertitletext{%
    \begin{center}
        \includegraphics[
            width=\textwidth
        ]{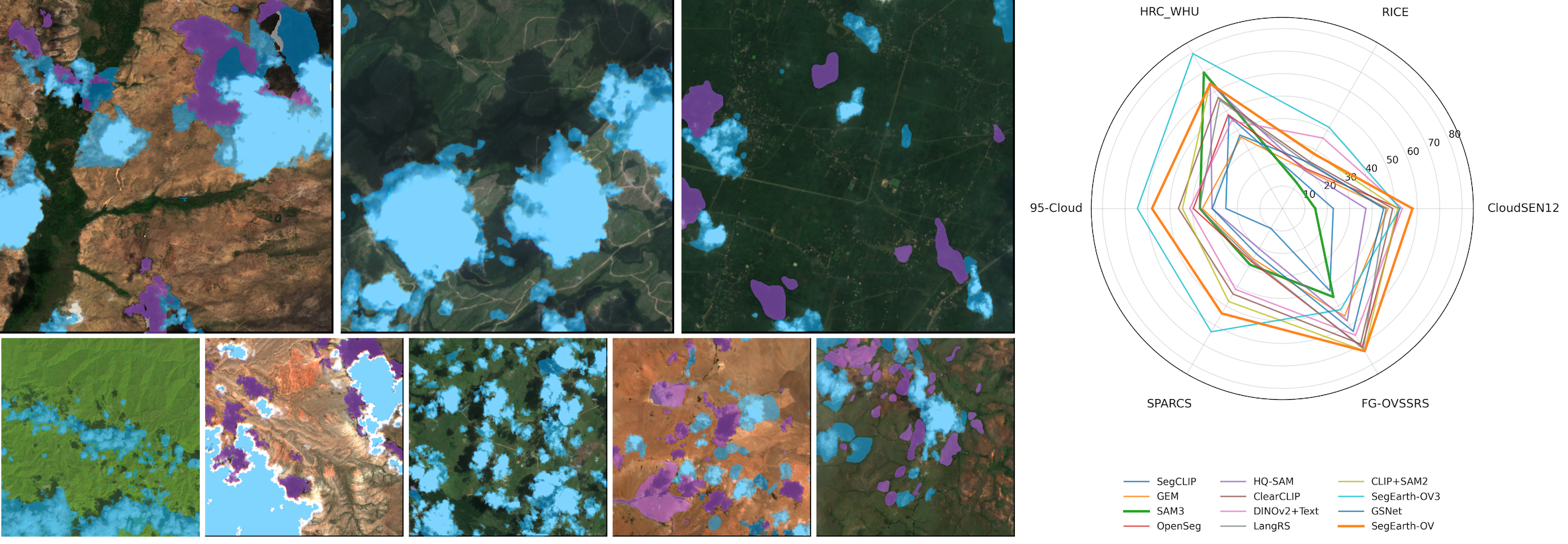}

        \scriptsize
        Overview of zero-shot open-vocabulary semantic segmentation (OVSS) capabilities for cloud detection in Earth Observation imagery. Left: Representative satellite images with overlaid cloud and cloud-shadow segmentation predictions, illustrating diverse cloud conditions and scene characteristics. Right: Comparison of 12 OVSS models across six cloud detection datasets using cumulative Intersection-over-Union (cIoU) for binary cloud segmentation. The results highlight variability in segmentation performance across models and datasets.
    \end{center}
    \vspace{0.3cm}
}

\begin{document}

\title{On the Capabilities and Limitations of Zero-Shot Open-Vocabulary Cloud Segmentation}

\author{
    \IEEEauthorblockN{
        Jose Sosa,\ %\orcidlink{},
        Thomas Gillet,\ %\orcidlink{},
        Rim Sleimi,\ %\orcidlink{},
        Albert Abelló\ %\orcidlink{}
    }
    \IEEEauthorblockA{
        Hydrosat, Luxembourg\\
        \{jsosa, tgillet, rsleimi, aabello\}@hydrosat.com
    }
}

\maketitle

\begin{abstract}
Recent advances in vision-language foundation models have enabled open-vocabulary semantic segmentation (OVSS) without task-specific training, offering new opportunities for Earth Observation (EO) applications. However, their capabilities for cloud detection remain insufficiently explored. In this work, we evaluate twelve OVSS approaches across six publicly available EO datasets in a zero-shot setting, i.e., without fine-tuning or adaptation to the target data. We investigate both binary cloud detection and multiclass segmentation, distinguishing between thick clouds, thin clouds, cloud shadows, and background. Our results show differences in performance across models and datasets, with SegEarth-OV3 achieving the highest binary cloud detection performance on four of the six datasets. While several approaches demonstrate promising performance in binary cloud detection, multiclass segmentation remains more challenging, particularly when distinguishing between fine-grained cloud categories. Furthermore, our controlled prompt sensitivity analysis reveals model-dependent responses to alternative textual descriptions, with appearance-oriented prompts benefiting certain approaches while degrading the performance of others. Overall, our findings demonstrate both the potential and current limitations of zero-shot OVSS for cloud detection, establishing a comparative benchmark for future research on generalisable cloud segmentation. Code will be available at \url{https://github.com/josesosajs/zero-shot-cloudovss}.
\end{abstract}


\section{Introduction}
Earth observation (EO) imagery is frequently affected by clouds and their shadows, which obscure the Earth's surface, alter the observed spectral information, and limit the operational use of EO data in various applications~\cite{ebel2020multisensor}. Therefore, accurate cloud detection remains an important preprocessing step for EO data analysis. Beyond binary cloud detection, distinguishing different cloud-related classes, such as thick clouds, thin clouds, haze, and cloud shadows, can enable downstream applications to apply class-specific filtering or correction strategies. However, developing robust models to distinguish these categories remains challenging~\cite{qiu2020cirrus}. In particular, lack of labelled resources, inconsistencies in annotation protocols, and class definitions across existing datasets make it difficult to establish large-scale, standardised training and evaluation resources~\cite{aybar2022cloudsen12}.

Early approaches to cloud detection predominantly relied on knowledge-driven methods based on spectral thresholds, physical properties, and manually designed decision rules to distinguish clouds from land, water, and other surface types~\cite{qiu2019fmask,louis2016sentinel,zekoll2021comparison,mahajan2020cloud}. Although these approaches remain widely used, complex atmospheric and surface conditions can make it difficult to delineate cloud boundaries and distinguish clouds from spectrally similar objects. With the development of deep learning, data-driven cloud detection methods have increasingly emerged as an alternative~\cite{jeppesen2019cloud,wang2024deep,li2021review,dong2024mcdnet,li2022gcdb}. In parallel, efforts by the EO community have resulted in several annotated cloud datasets, advancing the development of fully supervised learning approaches~\cite{aybar2022cloudsen12,aybar2024cloudsen12+}.

Despite this progress, supervised methods remain dependent on the availability and quality of annotated training data. This is particularly restrictive when extending cloud detection to new datasets, sensors, geographic regions, or more fine-grained cloud-related categories for which annotations may be scarce or defined differently~\cite{huang2023deep}. Recent advances in vision foundation models and open-vocabulary semantic segmentation (OVSS) provide a promising alternative. OVSS models establish links between visual and language representations, allowing semantic categories to be specified through textual class names or prompts at inference time~\cite{kombol2025survey,zhu2024survey}. Hence, these models recognise categories that were not explicitly indicated during task-specific training, making them attractive for annotation-scarce EO applications.

Previous research have explored OVSS with RGB EO imagery for specific tasks apart from standard multiclass segmentation, for instance, road detection, water-body detection, and building detection ~\cite{sosa2026enabling, li2025segearth, li2025segearthov3}. Nevertheless, cloud detection has received limited attention in the OVSS literature. Recent efforts have investigated fine-grained cloud segmentation~\cite{deng2026sano3} and highlighted the limitations of prompt engineering under domain shift, particularly for thin clouds and cloud shadows~\cite{kethavath2026low}. However, these investigations primarily focus on specific models or benchmarks, leaving the zero-shot capabilities and cross-dataset generalisation of current OVSS models insufficiently understood.

%Nevertheless, cloud detection has received limited attention in the related literature, with only a small number of studies considering clouds as semantic categories within their evaluations~\cite{deng2026sano3,kethavath2026low}. Consequently, the actual zero-shot capabilities of current OVSS models for cloud detection, and their ability to generalise across heterogeneous cloud datasets, remain insufficiently understood.

In this work, we present an evaluation of OVSS models for zero-shot cloud detection in EO imagery. We benchmark twelve publicly available approaches across six cloud detection datasets with varying levels of annotation granularity. All approaches are evaluated without task-specific training or adaptation on the target datasets. Our evaluation considers both binary cloud detection, which distinguishes clouds from the background, and multiclass cloud segmentation, which aims to differentiate between thick clouds, thin clouds, cloud shadows, and cloud-free regions. Furthermore, we investigate the sensitivity of OVSS models to variations in textual prompts, examining how different semantic descriptions and visual attributes influence segmentation performance. Through this study, we aim to provide a comprehensive assessment of the capabilities, generalisation, and limitations of current OVSS models for cloud detection. 

Our study makes the following contributions: \textbf{1) Zero-shot open-vocabulary cloud segmentation benchmarking.} We evaluate twelve OVSS approaches across six cloud segmentation datasets, providing a comparative assessment of their performance under different image characteristics. \textbf{2) Binary and multiclass cloud segmentation analysis.} We investigate the ability of OVSS models to distinguish clouds from the background and identify fine-grained categories, including thick clouds, thin clouds, and cloud shadows. All twelve models are evaluated for binary segmentation across six datasets, while a selected subset is evaluated for multiclass segmentation on the three datasets providing fine-grained annotations. \textbf{3) Textual prompt sensitivity analysis.} We examine how variations in textual inputs, including semantic, appearance-based, and descriptive formulations, affect cloud segmentation performance, assessing the robustness of OVSS models to small changes in prompt formulation.

\section{Related Work}

\noindent\textbf{Cloud detection.} Semantic segmentation plays a fundamental role in EO image understanding, supporting several applications like disaster management, urban planning, and environmental monitoring~\cite{spasev2023semantic}. Among these applications, cloud detection is particularly important, as clouds often obscure the Earth's surface, limiting the availability and reliability of optical satellite observations~\cite{ebel2020multisensor}. Early cloud detection approaches relied on handcrafted spectral features, threshold-based algorithms, and physical properties of clouds to distinguish cloudy from clear-sky regions. Examples include Fmask~\cite{qiu2019fmask} and Sen2Cor~\cite{louis2016sentinel}, which exploit spectral information and predefined decision rules to identify cloud-contaminated pixels. More recently, deep learning approaches, such as convolutional neural networks (CNNs) and transformer-based architectures, have demonstrated improved cloud segmentation capabilities by learning spatial and spectral representations directly from annotated remote sensing imagery~\cite{li2021review}. These methods have benefited from the availability of large-scale cloud detection datasets, e.g., CloudSEN12 and its extension, CloudSEN12+~\cite{aybar2022cloudsen12,aybar2024cloudsen12+}, which provide annotations for different cloud conditions, including thick clouds, thin clouds, and cloud shadows. Despite their effectiveness, most existing deep learning-based approaches operate within a closed-set segmentation framework, relying on predefined semantic categories and high-quality annotated data for training. Furthermore, variations in sensor characteristics, spatial resolution, atmospheric conditions, and geographical regions can limit their generalisation to previously unseen scenarios, often requiring domain-specific adaptation~\cite{kondylatos2025generalization}.

\noindent\textbf{OVSS.} Traditional semantic segmentation methods for EO data typically operate within a closed set of classes, requiring additional training or adaptation to recognise unseen categories. In contrast, OVSS offers a more flexible and scalable paradigm by segmenting novel semantic categories through natural language descriptions. Early OVSS approaches investigated the alignment of visual and textual representations within a shared embedding space~\cite{xian2019semantic,zhao2017open}. Subsequent research adopted CLIP~\cite{radford2021learning} as the core vision-language model for OVSS, leading to approaches such as SegCLIP~\cite{luo2023segclip}, ClearCLIP~\cite{lan2024clearclip}, GEM~\cite{bousselham2024grounding}, and OpenSeg~\cite{ghiasi2022scaling}, with applications extending to EO imagery~\cite{zhang2024rs5m,wang2024skyscript,liu2024remoteclip}. However, CLIP's limited spatial resolution and localisation capabilities present challenges for accurately delineating fine-grained structures. To address these limitations, recent approaches have incorporated complementary visual foundation models~\cite{sosa2026enabling, li2025segearthov3, li2025segearth}, such as DINOv2~\cite{jose2025dinov2}, SAM2~\cite{ravi2025sam}, and, more recently, SAM3~\cite{carion2026sam}, to improve spatial localisation and segmentation accuracy.

\begin{figure*}[h]
    \centering
    \includegraphics[width=\textwidth]{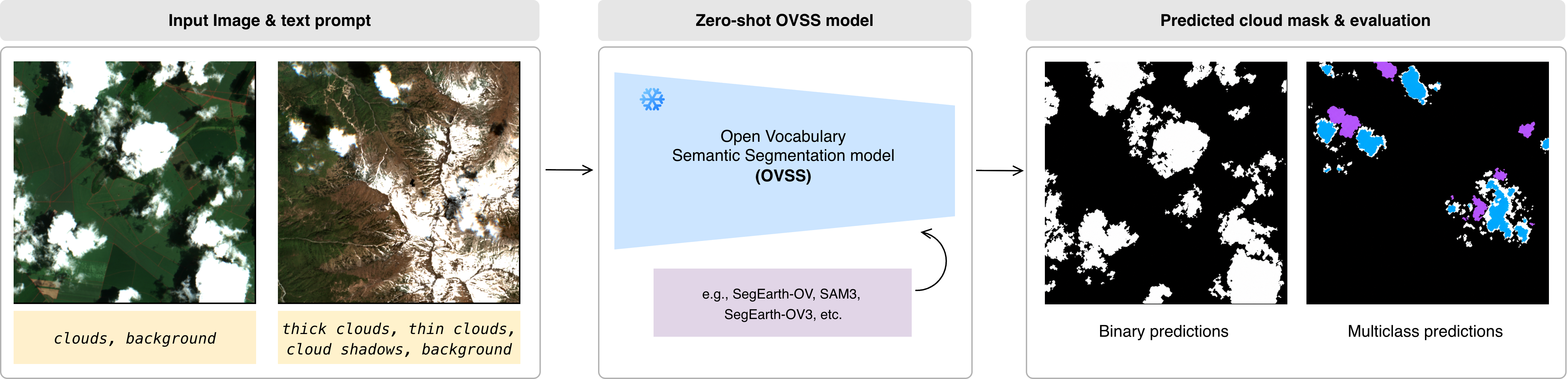}
    \caption{\textbf{Overview of the proposed zero-shot OVSS evaluation pipeline for cloud detection in EO imagery.} Input satellite images from six different datasets are processed together with natural-language prompts to perform binary or multiclass (when ground-truth annotations are available) segmentation. 12 different OVSS models, including SAM3, SegEarth-OV, and SegEarth-OV3, generate segmentation masks that distinguish between clouds and background in the binary segmentation setting, and among thick clouds, thin clouds, cloud shadows, and background in the multiclass setting. The predicted masks are subsequently evaluated against ground-truth annotations to quantitatively assess performance.}
    \label{fig:approach}
\end{figure*}

\begin{table*}[t]
\centering
\caption{\textbf{Summary of the cloud detection datasets used in our evaluation.}
$\checkmark$ indicates an explicitly annotated category, $\circ$ indicates general cloud annotations without thickness differentiation, and $\times$ indicates that the category is not separately annotated. Image counts correspond to the samples used for evaluation.}
\label{tab:datasets}
\resizebox{\textwidth}{!}{
\begin{tabular}{lcccccc}
\hline
\textbf{Dataset} &
\textbf{Source} &
\textbf{Main region} &
\multicolumn{3}{c}{\textbf{Cloud annotations}} &
\textbf{Number of images} \\
\cline{4-6}
& & &
\textbf{Thick cloud} &
\textbf{Thin cloud} &
\textbf{Cloud shadow} &
\\
\hline

CloudSEN12~\cite{aybar2022cloudsen12}
& S2 & Global & $\checkmark$ & $\checkmark$ & $\checkmark$ & 2450 \\

FG-OVSSRS~\cite{deng2026sano3}
& S2 & Global & $\checkmark$ & $\checkmark$ & $\checkmark$ & 1954 \\

SPARCS~\cite{hughes2014automated}
& L8 & Global & $\circ$ & $\circ$& \checkmark & 80 \\

RICE~\cite{lin2019remote}
& L8 & Global & $\circ$ & $\circ$ & $\times$ & 450 \\

HRC\_WHU~\cite{li2019deep}
& Google Earth & Global & $\circ$& $\circ$ & $\times$ & 150 \\

95-Cloud~\cite{mohajerani2021cloud}
& L8 & USA & $\circ$ & $\circ$ & $\times$ & 916 \\

\hline
\end{tabular}}
\end{table*}

Within the EO domain, existing OVSS approaches differ in their use of domain-specific adaptation, architectural components, and training strategies. Methods such as GSNet~\cite{ye2025towards} adopt CLIP's image and text encoders and couple them with a remote sensing specialist encoder, which acts as a domain prior to enhance CLIP capabilities on remote sensing data. In contrast, other approaches seek to minimise the dependence on domain-specific training. For instance, SegEarth-OV~\cite{li2025segearth} enhances CLIP's spatial representations through SimFeatUp (based on FeatUp~\cite{fu2024featup}), a feature upsampling module pre-trained on remote sensing data to improve fine-grained segmentation. Other approaches combine vision-language models with promptable segmentation architectures, such as LangRS~\cite{DIAB2025100105}, which integrates Grounding DINO~\cite{liu2024grounding} and SAM~\cite{kirillov2023segment} for text-guided segmentation. Similarly, Sosa et al.~\cite{sosa2026enabling} combine CLIP with SAM2~\cite{ravi2025sam}, using the latter to generate class-agnostic segmentation masks that are subsequently assigned semantic labels through vision-language alignment. Unlike SegEarth-OV, this approach does not require EO-specific training, enabling fully zero-shot inference. More recently, the introduction of SAM3 has motivated approaches such as SegEarth-OV3~\cite{li2025segearthov3}, which exploits its text-conditioned segmentation capabilities to boost performance on remote sensing benchmarks without requiring additional domain-specific training.

\noindent\textbf{OVSS for cloud detection.} Despite recent advances in OVSS for EO data, its application to cloud detection remains underexplored compared with other applications, such as building detection, water bodies detection, and road detection, which are normally included within evaluation benchmarks~\cite{sosa2026enabling,li2025segearth, li2025segearthov3}. Unlike conventional cloud detection models trained on task-specific annotations, OVSS approaches offer the potential to identify clouds and related categories directly from textual descriptions. However, cloud segmentation presents particular challenges for general-purpose segmentation models, including the variable appearance of clouds, ambiguous boundaries between thin clouds and clear-sky regions, and the visual similarity between clouds and bright surfaces such as snow, water, or urban structures. These challenges, together with the limited spectral information available in RGB imagery compared with multispectral observations, raise questions regarding the ability of existing OVSS methods to distinguish cloud categories and generalise across satellite imagery acquired under diverse conditions. Motivated by some of these limitations, this work investigates the zero-shot cloud segmentation capabilities of recent OVSS approaches across several datasets supporting cloud detection, examining their performance under both binary and multiclass segmentation settings.

\section{Approach}

We evaluate twelve OVSS approaches ~\cite{luo2023segclip,ke2023segment,lan2024clearclip,DIAB2025100105,bousselham2024grounding,ghiasi2022scaling,jose2025dinov2,sosa2026enabling,carion2026sam,li2025segearthov3,li2025segearth,ye2025towards} for the task of cloud detection across six datasets~\cite{aybar2022cloudsen12, deng2026sano3, hughes2014automated, lin2019remote, li2019deep, mohajerani2021cloud}. Our evaluation pipeline is illustrated in \autoref{fig:approach}. To ensure consistency across methods and datasets, all input images are restricted to RGB channels, taken from dataset-provided composites or the corresponding visible bands. Although this excludes potentially informative spectral bands, it enables a controlled comparison under a common input configuration compatible with the requirements of the selected OVSS models. All approaches are evaluated without fine-tuning or parameter adaptation on the target cloud datasets. Models differ, however, in whether their pretrained components incorporate EO imagery. For instance, of all the evaluated methods, ten were trained on general-domain imagery~\cite{luo2023segclip,ke2023segment,lan2024clearclip,DIAB2025100105,bousselham2024grounding,ghiasi2022scaling,jose2025dinov2,sosa2026enabling,carion2026sam,li2025segearthov3}, while two are trained either entirely~\cite{ye2025towards} or partially~\cite{li2025segearth} using remote sensing data.

Our main experiments involve two segmentation scenarios: binary and multiclass. In the binary setting, all models are evaluated across the six datasets to distinguish clouds from background, using the textual prompts \textit{``clouds''} and \textit{``background''}, respectively, unless otherwise specified. For the multiclass setting, we select the top three overall performing models from the binary evaluation (SegEarth-OV3, SegEarth-OV, and CLIP+SAM2) along with SAM3 (the foundational base model for SegEarth-OV3). The multiclass evaluation uses CloudSEN12, FG-OVSSRS, and SPARCS datasets. We consider four classes using the textual prompts \textit{``thick clouds''}, \textit{``thin clouds''}, \textit{``cloud shadows''}, and \textit{``background''}. Since the SPARCS dataset does not distinguish cloud thickness in its annotations, we use three classes defined by the prompts \textit{``clouds''}, \textit{``cloud shadows''}, and \textit{``background''}.

\noindent\textbf{Data.} Our evaluation considers six diverse, publicly available datasets containing human-annotated cloud masks~\cite{aybar2022cloudsen12, hughes2014automated, lin2019remote, li2019deep, mohajerani2021cloud, deng2026sano3}. The selected datasets cover different sensors, spatial resolutions, geographical regions, and cloud conditions, providing a heterogeneous benchmark for assessing the zero-shot generalisation of OVSS models. Table~\ref{tab:datasets} summarises their characteristics and the number of images used for evaluation. We use the complete SPARCS and HRC\_WHU datasets, the full RICE2 subset of RICE, and the cloud category of FG-OVSSRS~\cite{deng2026sano3}. For CloudSEN12 and 95-Cloud, we evaluate randomly selected subsets of 2,450 and 916 images, respectively. In the case of FG-OVSSRS, we manually inspected the images and excluded samples containing large no-data regions, which were frequently observed in this dataset. For binary evaluation, thick and thin clouds are merged into a single positive class in datasets with these fine-grained annotations (CloudSEN12 and FG-OVSSRS), while all remaining categories are assigned to the background. As shown in \autoref{tab:datasets}, only CloudSEN12, FG-OVSSRS, and SPARCS provide the class-level annotations required for quantitative multiclass evaluation. Accordingly, we conduct the multiclass evaluation on these three datasets. The exact sample lists and evaluation configurations are provided in our public repository to ensure reproducibility.

%The exact sample lists and evaluation configurations are provided in our public repository to ensure reproducibility. 

% A summary of the datasets is provided in Table~\ref{tab:datasets}. For datasets evaluated using randomly selected subsets, the exact sample lists are made available in our public repository to ensure reproducibility. Note that, for FG-OVSSRS~\cite{deng2026sano3}, we use the data from the cloud category.

% Dataset image cell
\newcommand{\datasetcell}[2]{%
    \begin{minipage}[t]{0.48\columnwidth}
        \centering
        \parbox[c][2.2cm][c]{\linewidth}{%
            \centering
            \includegraphics[
                width=\linewidth,
                height=2.2cm,
                keepaspectratio
            ]{#2}%
        }
        \par\vspace{2pt}
        {\scriptsize\textbf{#1}}
    \end{minipage}%
}

% Dataset row: name + three representative images
\newcommand{\datasetrow}[4]{%
    \noindent{\scriptsize\textbf{#1}}\par\vspace{2pt}
    \begin{minipage}{0.32\columnwidth}
        \centering
        \includegraphics[
            width=\linewidth,
            height=1.6cm,
            keepaspectratio
        ]{#2}
    \end{minipage}%
    \hfill
    \begin{minipage}{0.32\columnwidth}
        \centering
        \includegraphics[
            width=\linewidth,
            height=1.6cm,
            keepaspectratio
        ]{#3}
    \end{minipage}%
    \hfill
    \begin{minipage}{0.32\columnwidth}
        \centering
        \includegraphics[
            width=\linewidth,
            height=1.6cm,
            keepaspectratio
        ]{#4}
    \end{minipage}%
    \par\vspace{5pt}
}

\noindent\textbf{Evaluation metrics}. In line with previous work~\cite{r1-li2025segearth}, the predicted segmentation masks are evaluated against their corresponding ground-truth annotations using two standard metrics, global IoU (gIoU) and cumulative IoU (cIoU). The gIoU is computed independently for each image and subsequently averaged across the dataset, whereas cIoU is calculated by accumulating the intersections and unions across all images before computing their ratio:

\[
\mathrm{gIoU}_c = \frac{1}{|\mathcal{V}_c|} \sum_{i \in \mathcal{V}_c} \frac{|\hat{Y}_{i,c} \cap Y_{i,c}|}{|\hat{Y}_{i,c} \cup Y_{i,c}|},
\]

\[
\mathrm{cIoU}_c = \frac{\sum_{i=1}^{N} |\hat{Y}_{i,c} \cap Y_{i,c}|}{\sum_{i=1}^{N} |\hat{Y}_{i,c} \cup Y_{i,c}|},
\]

\noindent where \(N\) denotes the number of images, \(\hat{Y}_{i,c}\) and \(Y_{i,c}\) represent the predicted and ground-truth pixel sets for class \(c\) in image \(i\), respectively, and \(\mathcal{V}_c\) denotes the subset of images with a nonempty union for class \(c\). Images with an empty union are excluded from the corresponding class-wise gIoU average. Consequently, correctly predicted class absence does not contribute to gIoU, while false-positive predictions on images where the class is absent receive an IoU of zero.

While gIoU assigns equal importance to each valid image regardless of cloud coverage, cIoU provides a dataset-level measure that gives greater weight to images with larger cloud regions. For binary segmentation, both metrics are reported only for the cloud class. For multiclass segmentation, we report class-wise gIoU and its average across all classes, including background.

\noindent\textbf{Implementation details.} Whenever available, we use the official implementations and publicly released pretrained weights, introducing only the modifications necessary to integrate the evaluated datasets into our pipeline. All models operate on RGB images, following their respective input preprocessing requirements. Unless otherwise specified, we retain the inference configurations provided by the original implementations. To ensure consistent evaluation, predictions from the different approaches are converted into discrete semantic label maps and resized to the corresponding ground-truth dimensions using nearest-neighbour interpolation. Model-specific postprocessing procedures, including mask aggregation, class assignment, and the handling of overlapping predictions, follow their respective implementations. Inference for all experiments is conducted using GPU acceleration.

\definecolor{avgbg}{RGB}{245,245,250}

\begin{table*}[t]
\centering
\caption{\textbf{Quantitative evaluation of binary cloud segmentation.} gIoU and cIoU scores for the cloud class across six cloud segmentation datasets. The best results are highlighted in \textbf{bold}, while the second-best results are \underline{underlined}.}
\label{tab:iou_comparison}

\setlength{\tabcolsep}{4pt}
\renewcommand{\arraystretch}{1.3}
\small

\resizebox{\textwidth}{!}{%
\begin{tabular}{@{}lcccccccccccc|cc@{}}
\toprule

\textbf{Model}
& \multicolumn{2}{c}{\textbf{CloudSEN12}}
& \multicolumn{2}{c}{\textbf{RICE}}
& \multicolumn{2}{c}{\textbf{HRC\_WHU}}
& \multicolumn{2}{c}{\textbf{95-Cloud}}
& \multicolumn{2}{c}{\textbf{SPARCS}}
& \multicolumn{2}{c}{\textbf{FG-OVSSRS}}
& \multicolumn{2}{c}{\textbf{Average}} \\

\cmidrule(lr){2-3}
\cmidrule(lr){4-5}
\cmidrule(lr){6-7}
\cmidrule(lr){8-9}
\cmidrule(lr){10-11}
\cmidrule(lr){12-13}
\cmidrule(lr){14-15}

& gIoU & cIoU
& gIoU & cIoU
& gIoU & cIoU
& gIoU & cIoU
& gIoU & cIoU
& gIoU & cIoU
& gIoU & cIoU \\
\midrule

\rowcolor{groupgray}
\multicolumn{15}{l}{\textit{Without RS-specific training}} \\

SegCLIP \textsubscript{\scriptsize\textcolor{venuegray}{ICML'23}}~\cite{luo2023segclip}
& 15.81	& 22.67
& 8.46 & 15.22
& 42.60 & 47.48
& 19.91 & 25.14
& 5.24 & 10.22
& 30.96 & 42.24
& 20.50	& 27.16 \\

GEM \textsubscript{\scriptsize\textcolor{venuegray}{CVPR'24}}~\cite{bousselham2024grounding}
& 46.26	& 47.00
& 20.00 & 20.02
& 36.86 & 36.68
& 33.18 & 35.54
& 20.78 & 25.86
& 47.43	& 55.31
& 34.08	& 36.74 \\

SAM3 \textsubscript{\scriptsize\textcolor{venuegray}{ICLR'26}}~\cite{carion2026sam}
& 20.14	& 14.62
& 10.05 & 13.03
& \underline{67.85} & \underline{69.82}
& 39.56 & 36.66
& 28.77 & 28.81
& 38.61	& 45.42
& 34.16	& 34.73 \\

OpenSeg \textsubscript{\scriptsize\textcolor{venuegray}{ECCV'22}}~\cite{ghiasi2022scaling}
& 44.14	& 52.12
& 15.84 & 20.59
& 50.15 & 48.12
& 31.49 & 39.59
& 17.93 & 27.27
& \underline{63.10}	& 71.08
& 37.11	& 43.13 \\

HQ-SAM \textsubscript{\scriptsize\textcolor{venuegray}{NeurIPS'23}}~\cite{ke2023segment}
& 38.01	& 37.18
& 19.26 & 19.63
& 66.06 & 64.17
& 32.79 & 31.27
& 21.43 & 20.67
& 54.38	& 57.69
& 38.65	& 38.43 \\

ClearCLIP \textsubscript{\scriptsize\textcolor{venuegray}{ECCV'24}}~\cite{lan2024clearclip}
& 44.57	& 49.09
& 15.19 & 24.15
& 52.70 & 57.00
& 42.41 & 46.21
& 32.70 & 43.91
& 55.05	& 69.61
& 40.43 & 48.33 \\

DINOv2$+$Text \textsubscript{\scriptsize\textcolor{venuegray}{CVPR'25}}~\cite{jose2025dinov2}
& 49.28	& \underline{53.36}
& 23.89 & \underline{36.18}
& 43.40 & 45.08
& 39.30 & 41.19
& 34.41 & 41.54
& 53.55	& 65.10
& 40.64	& 47.07 \\

LangRS \textsubscript{\scriptsize\textcolor{venuegray}{AIGS'25}}~\cite{DIAB2025100105}
& 44.78	& 46.09
& 22.30 & 23.05
& 57.74 & 55.69
& 36.97 & 36.90
& 27.50 & 26.61
& \textbf{63.30} & 72.05
& 42.10	& 43.40 \\

CLIP$+$SAM2 \textsubscript{\scriptsize\textcolor{venuegray}{CVPRw'26}}~\cite{sosa2026enabling}
& 45.53	& 51.34
& 20.96 & 28.44
& 65.12 & 64.34
& 44.02 & 44.44
& 41.95 & 47.76
& 60.46 & \underline{73.10}
& 46.34	& 51.57 \\

SegEarth-OV3 \textsubscript{\scriptsize\textcolor{venuegray}{arXiv'25}}~\cite{li2025segearthov3}
& \underline{53.67} & 52.38
& \textbf{31.88} & \textbf{41.67}
& \textbf{76.88} & \textbf{79.54}
& \textbf{63.06} & \textbf{64.49}
& \textbf{60.68} & \textbf{63.31}
& 47.11	& 51.93
& \textbf{55.55} & \textbf{58.89} \\

\midrule
\rowcolor{groupgray}
\multicolumn{15}{l}{\textit{With RS-specific training}} \\

GSNet \textsubscript{\scriptsize\textcolor{venuegray}{AAAI'25}}~\cite{ye2025towards}
& 44.73	& 45.07
& 19.67 & 23.57
& 37.43 & 37.74
& 31.02 & 31.20
& 21.42 & 23.24
& 62.27	& 63.09
& 36.09	& 37.32 \\

SegEarth-OV \textsubscript{\scriptsize\textcolor{venuegray}{CVPR'25}}~\cite{li2025segearth}
& \textbf{55.25} & \textbf{	57.90}
& \underline{24.06} & 28.07
& 66.34 & 64.18
& \underline{57.29} & \underline{57.98}
& \underline{44.80} & \underline{53.89}
& 62.76	& \textbf{73.27}
& \underline{51.75} & \underline{55.88} \\

\bottomrule
\end{tabular}%
}
\end{table*}

\begin{table*}[t]
\centering
\caption{\textbf{Quantitative evaluation of multiclass cloud segmentation.} gIoU scores for each semantic class across the CloudSEN12, FG-OVSSRS, and SPARCS datasets. The Avg. columns report the average scores across four classes for CloudSEN12 and FG-OVSSRS, and three classes for SPARCS, including background. The best results are highlighted in \textbf{bold}, while the second-best results are \underline{underlined}.}
\label{tab:class_iou_results}

\setlength{\tabcolsep}{6pt}
\renewcommand{\arraystretch}{1.3}

\resizebox{\textwidth}{!}{
\begin{tabular}{lcccc|c|cccc|c|ccc|c}
\toprule
& \multicolumn{5}{c}{\textbf{CloudSEN12}}
& \multicolumn{5}{c}{\textbf{FG-OVSSRS}}
& \multicolumn{4}{c}{\textbf{SPARCS}} \\

%\cmidrule(lr){2-6}
%\cmidrule(lr){7-11}
%\cmidrule(lr){12-15}

\textbf{Model}
& Thick C.
& Thin C.
& C. Shadow
& Back.
& \textbf{Avg.}
& Thick C.
& Thin C.
& C. Shadow
& Back.
& \textbf{Avg.}
& Cloud
& C. Shadow
& Back.
& \textbf{Avg.} \\
\midrule

CLIP + SAM2~\cite{sosa2026enabling}
& 39.53 & 3.45 & 11.63 & 38.52 & \underline{23.28}
& 42.15 & 0.63 & 0.36 & 35.26 & 19.60
& 30.98 & 4.14 & 64.50 & 33.21 \\

SAM3~\cite{carion2026sam}
& 8.50 & 0.01 & 0.13 & 45.16 & 13.45
& 45.96 & 1.89 & 0.00 & 35.90 & \underline{20.94}
& 28.77	& 0.00	& 76.89	& 35.22 \\

SegEarth-OV~\cite{li2025segearth}
& 44.85 & 2.27 & 6.64 & 51.84 & \textbf{26.40}
& 46.52 & 6.34 & 0.11 & 31.65 & \textbf{21.16}
& 45.06 & 1.70 & 78.52 & \underline{41.76} \\

SegEarth-OV3~\cite{li2025segearthov3}
& 15.90 & 0.33 & 5.19 & 53.33 & 18.69
& 44.66 & 4.75 & 0.00 & 34.18 & 20.90
& 61.00	& 2.06 & 82.07	& \textbf{48.38} \\

\bottomrule
\end{tabular}%
}
\end{table*}

% \begin{table*}[t]
% \centering
% \caption{\textbf{Quantitative evaluation of multiclass cloud segmentation.} Average per-image IoU scores for each semantic class across the CloudSEN12 and FG-OVSSRS datasets. The mIoU columns report the mean IoU across all four classes, including background. The best results are highlighted in \textbf{bold}, while the second-best results are \underline{underlined}.}
% \label{tab:class_iou_results}
% \setlength{\tabcolsep}{6pt}
% \renewcommand{\arraystretch}{1.3}

% \resizebox{\textwidth}{!}{
% \begin{tabular}{lcccc|c|cccc|c}
% \toprule
% & \multicolumn{5}{c}{\textbf{CloudSEN12}}
% & \multicolumn{5}{c}{\textbf{FG-OVSSRS}} \\

% \cmidrule(lr){2-6}
% \cmidrule(lr){7-11}

% \textbf{Model}
% & Thick cloud
% & Thin cloud
% & Cloud shadow
% & Back.
% & \textbf{mIoU}
% & Thick cloud
% & Thin cloud
% & Cloud shadow
% & Back.
% & \textbf{mIoU} \\
% \midrule

% CLIP + SAM2~\cite{sosa2026enabling}
% & 39.53	 &	3.45	&	11.63	&	38.52	&	\underline{23.28}
% &42.15	&	0.63	&0.36	&35.26&	19.60\\

% SAM3~\cite{carion2026sam}
% & 8.50	&	0.01	&	0.13	&	45.16	&	13.45
% & 45.96	&	1.89	&0.00	&35.90	& \underline{20.94} \\

% SegEarth-OV~\cite{li2025segearth}
% & 44.85	& 2.27 &		6.64		& 51.84 &		\textbf{26.40}
% & 46.52 & 6.34 &	0.11 &	31.65 &	\textbf{21.16} \\
% %& 53.44 & 0.91 & 7.30 & 52.44 & \underline{28.52}
% %& 47.65	& 0.87 & 0.28 &	33.05 &	\underline{20.46} \\

% SegEarth-OV3~\cite{li2025segearthov3}
% & 15.90	&	0.33	&	5.19	&	53.33	&	18.69
% & 44.66	& 4.75	& 0.00	& 34.18 &	20.90 \\

% \bottomrule
% \end{tabular}%
% }
% \end{table*}

\section{Experiments and results}

\noindent\textbf{Binary cloud segmentation.} Table \ref{tab:iou_comparison} presents the zero-shot performance of OVSS models across the six cloud segmentation datasets in terms of gIoU and cIoU. Note that these scores only represent the cloud class. Overall, SegEarth-OV3~\cite{li2025segearthov3} achieves the highest average performance, obtaining the best results on four of the six datasets.

Approaches incorporating EO-specific training or adaptation, such as GSNet~\cite{ye2025towards} and SegEarth-OV~\cite{li2025segearth}, show varying performance relative to general-purpose counterparts. While SegEarth-OV achieves higher scores than SegEarth-OV3 on CloudSEN12 and FG-OVSSRS, SegEarth-OV3 performs better on the remaining four datasets. Complementing these quantitative results, \autoref{fig:qualitative_comparison_1} depicts qualitative comparisons of binary cloud segmentation across representative models and samples from four of the datasets.

Given that SegEarth-OV3~\cite{li2025segearthov3} builds upon SAM3~\cite{carion2026sam}, one might expect SAM3~\cite{carion2026sam} itself to exhibit strong segmentation capabilities. However, based on our results, SAM3~\cite{carion2026sam} alone achieves limited performance across most datasets. This difference may be partly explained by the additional architectural and inference components introduced in SegEarth-OV3, including mask fusion and presence-score filtering, although their individual contributions are not isolated in our evaluation. Beyond model architecture, textual prompt formulation may also influence segmentation results. In particular, SAM3 may benefit from more descriptive prompts that characterise the visual appearance of the target objects, rather than the simple class names used in our baseline evaluation~\cite{carion2026sam}. We therefore investigate the sensitivity of representative models to alternative textual descriptions in a subsequent analysis.

\noindent\textbf{Multiclass cloud segmentation.} In addition to binary cloud detection, we evaluate four OVSS models in a multiclass setting. Table \ref{tab:class_iou_results} presents the class-wise performance on the CloudSEN12, FG-OVSSRS, and SPARCS datasets. SegEarth-OV achieves the highest average scores on CloudSEN12 and FG-OVSSRS, while SegEarth-OV3 performs best on SPARCS. Interestingly, despite achieving the highest average performance in binary cloud detection, SegEarth-OV3 does not consistently outperform the other approaches in the multiclass setting. This performance difference highlights that strong binary cloud detection performance does not necessarily translate into effective discrimination between cloud categories.

Across the three evaluated datasets, the models demonstrate stronger performance on thick clouds and background than on thin clouds and cloud shadows. For example, on CloudSEN12, thin-cloud score remains below 3.5 for all four models, while thick-cloud score reaches 44.85 for SegEarth-OV. Similarly, on SPARCS, SegEarth-OV3 achieves an average of 48.38, despite obtaining only 2.06 for cloud shadows.

\begin{figure}
    \centering
    \includegraphics[width=\linewidth]{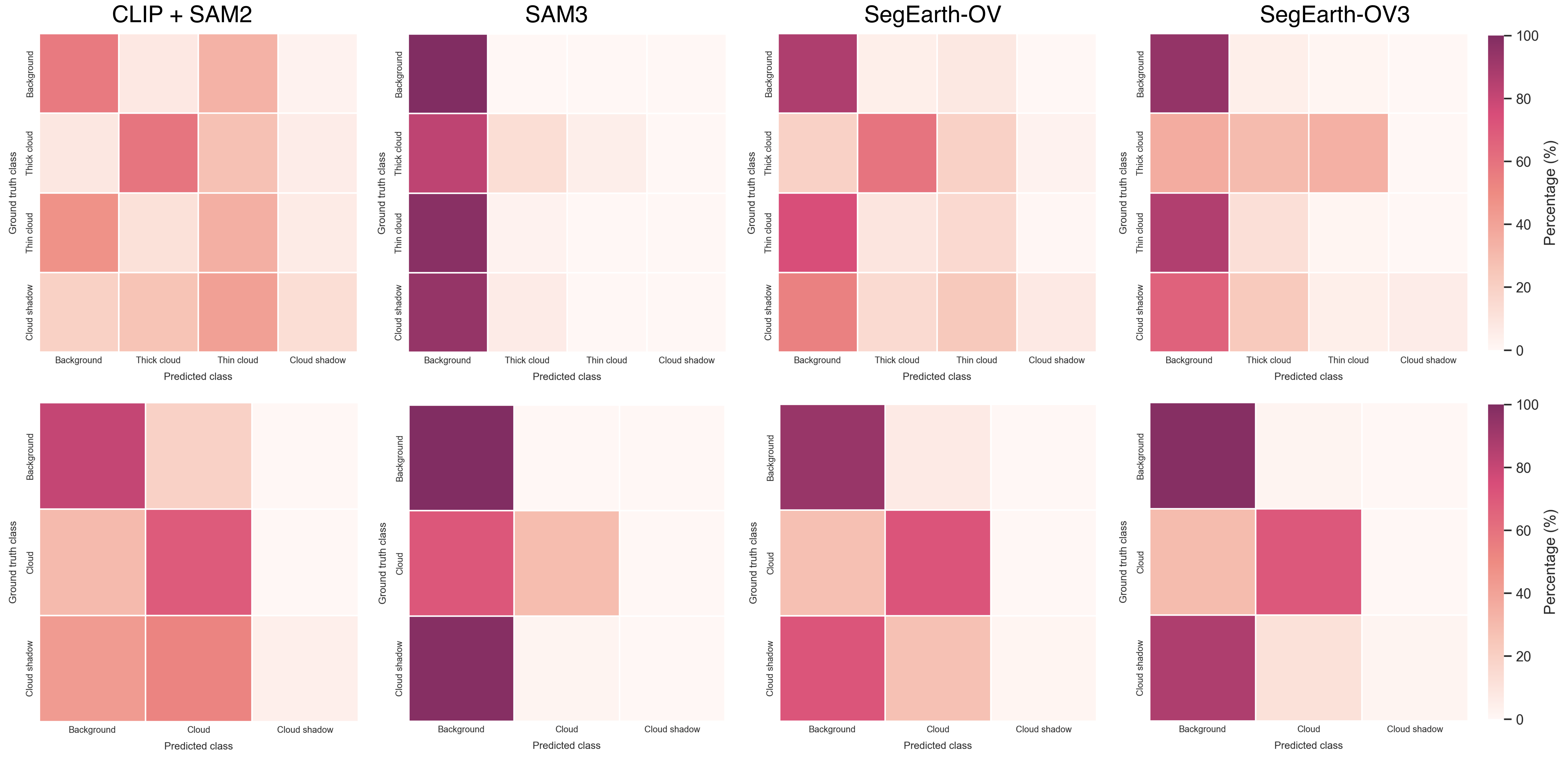}
    \caption{Row-normalised confusion matrices for multiclass cloud segmentation on CloudSEN12 (first row) and SPARCS (second row) datasets. Rows represent ground-truth classes and columns represent predicted classes. Values indicate the percentage of pixels assigned to each predicted class, with each row normalised independently. The matrices highlight model-specific misclassification patterns.}
    \label{fig:multiclass_confusion_matrices}
\end{figure}

The confusion matrices for CloudSEN12 and SPARCS, depicted in~\autoref{fig:multiclass_confusion_matrices}, provide further insight into models' limitations. On CloudSEN12, SAM3 exhibits a strong bias toward predicting background, often misclassifying all three cloud-related categories as background. In contrast, SegEarth-OV demonstrates better discrimination of thick clouds, although thin clouds and cloud shadows remain confused with background. CLIP+SAM2 exhibits a different error pattern, tending to overpredict thin clouds, while SegEarth-OV3 shows confusion between cloud categories and background. Similar limitations are observed on SPARCS, where cloud shadows are frequently misclassified as background or clouds. Confusion matrices for FG-OVSSRS are omitted due to space constraints. 

Overall, the multiclass results show that, although zero-shot OVSS models can identify cloud regions, as in the binary setting, distinguishing fine-grained cloud categories remains challenging. This limitation may be related to the overlapping visual characteristics of cloud categories, differences in their representation within pretrained models, and inconsistencies in annotation criteria across datasets~\cite{lopez2021benchmarking,aybar2022cloudsen12,aybar2024cloudsen12+}.

\noindent\textbf{Prompt sensitivity analysis.} Beyond model architecture and domain-specific training, the choice of textual class descriptions can influence the performance of OVSS models. However, the impact of prompt formulation has received minor attention in evaluations of OVSS models with EO data~\cite{park2025personalized,zhu2024survey,kethavath2026low}. To investigate this aspect, we conduct a controlled prompt sensitivity analysis using a predefined set of textual formulations designed to capture different ways of describing cloud-related categories.

We focus on the four approaches evaluated in the multiclass setting, i.e., CLIP+SAM2~\cite{sosa2026enabling}, SAM3~\cite{carion2026sam}, SegEarth-OV~\cite{li2025segearth}, and SegEarth-OV3~\cite{li2025segearthov3}. For binary segmentation, we evaluate each prompt configuration across all six datasets and report the average performance. For multiclass segmentation we restrict the prompt sensitivity analysis to CloudSEN12 and FG-OVSSRS datasets to ensure consistency in the annotated classes. We consider five prompt configurations:

\begin{table}[]
\centering
\caption{\textbf{Textual prompt configurations used in the sensitivity analysis.} The background prompt remains unchanged across all
configurations.}
\label{tab:prompt_configurations}

\setlength{\tabcolsep}{5pt}
\renewcommand{\arraystretch}{1.15}
\footnotesize

\resizebox{\columnwidth}{!}{%
\begin{tabular}{@{}ll@{}}
\toprule
\textbf{Prompt Category} & \textbf{Textual Prompts} \\
\midrule

\rowcolor{groupgray}\multicolumn{2}{l}{\textit{Binary segmentation}} \\
Baseline     & clouds \\
Semantic     & cloud cover \\
Appearance   & white clouds \\
Descriptive  & white atmospheric clouds \\
Visual Proxy & white areas \\

\midrule
\rowcolor{groupgray}\multicolumn{2}{l}{\textit{Multiclass segmentation}} \\
Baseline     & thick clouds, thin clouds, cloud shadows \\
Semantic     & dense clouds, wispy clouds, cloud-cast shadows \\
Appearance   & opaque clouds, translucent clouds, dark shadows \\
Descriptive  & dense white clouds, thin transparent clouds, \\
             & dark cloud shadows \\
Visual Proxy & white areas, hazy areas, dark areas \\

\bottomrule
\end{tabular}%
}%
\end{table}
 
\begin{enumerate}
    \item \textit{Baseline:} Uses the original class names from our main experiments.
    \item \textit{Semantic:} Employs alternative terms with similar semantic meanings.
    \item  \textit{Appearance:} Emphasises the visual characteristics of the target classes.
    \item \textit{Descriptive:} Combines semantic and visual attributes into more detailed descriptions.
    \item  \textit{Visual Proxy:} Replaces semantic class labels with descriptions of visually observable features that may serve as proxies for the target categories.
\end{enumerate}

% Single example row: Image + GT + 8 models
\newcommand{\visrowrice}{%
    \viscell{{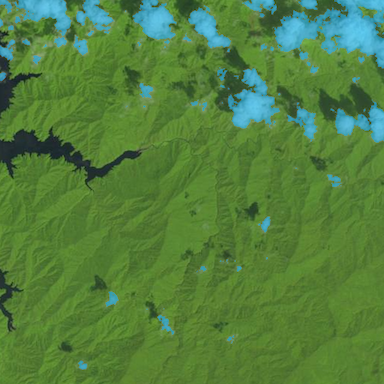}}
    & \viscell{{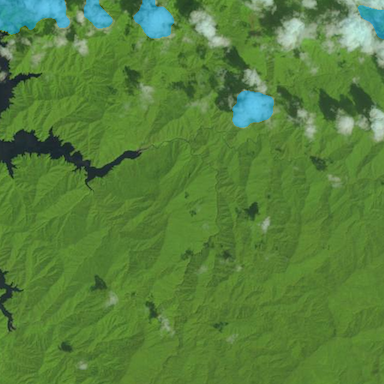}}
    & \viscell{{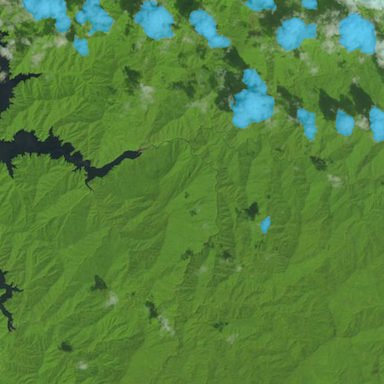}}
    & \viscell{{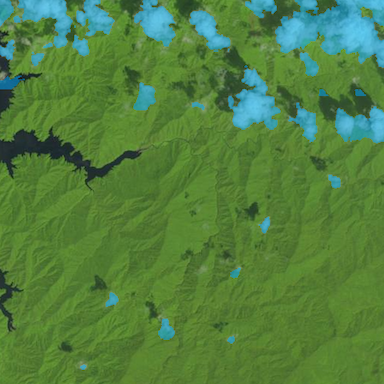}}
    & \viscell{{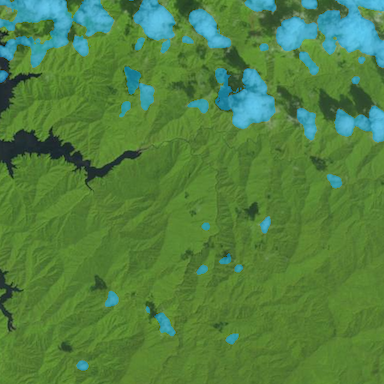}}
    & \viscell{{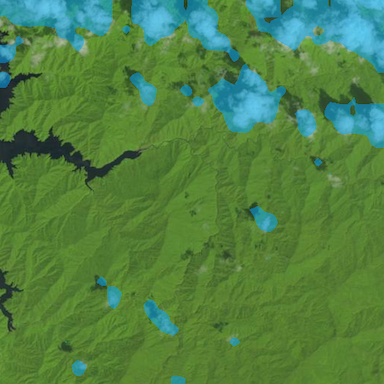}}
    & \viscell{{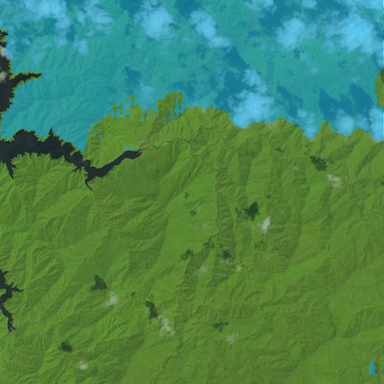}}
    & \viscell{{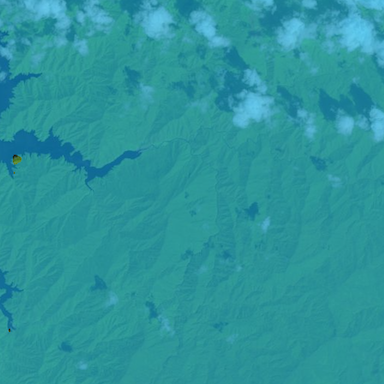}}
    & \viscell{{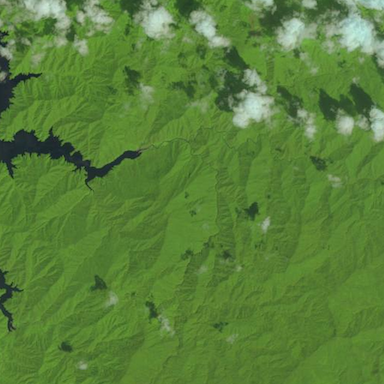}}
    & \viscell{{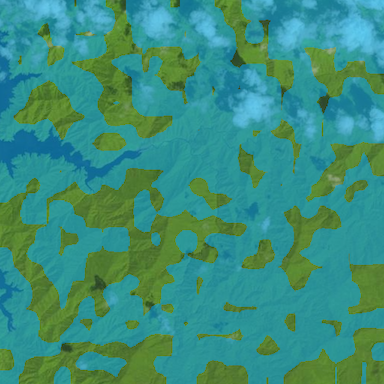}}%
}

\newcommand{\visrowcloud}{%
    \viscell{{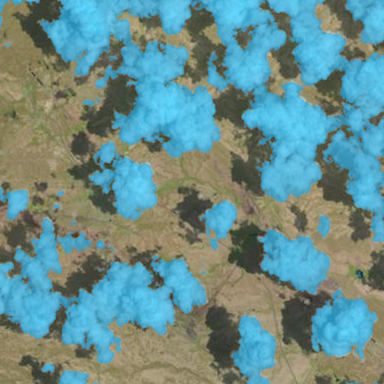}}
    & \viscell{{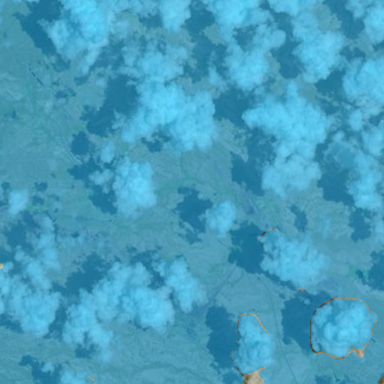}}
    & \viscell{{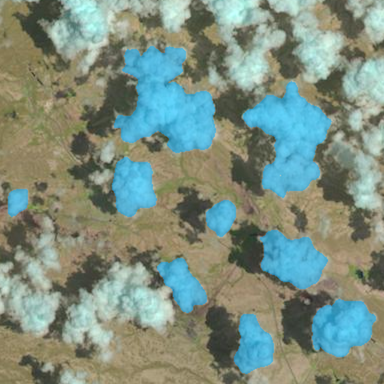}}
    & \viscell{{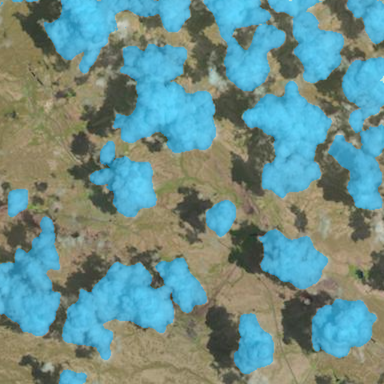}}
    & \viscell{{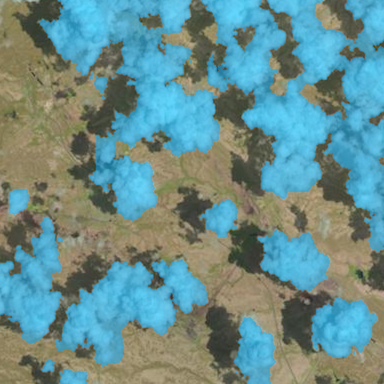}}
    & \viscell{{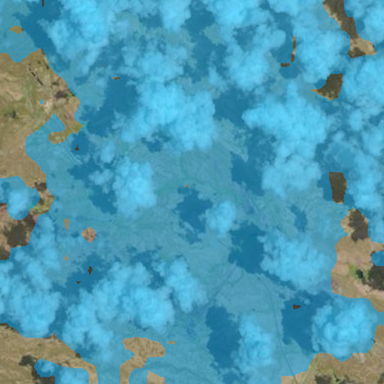}}
    & \viscell{{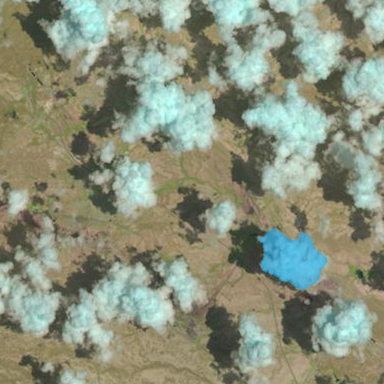}}
    & \viscell{{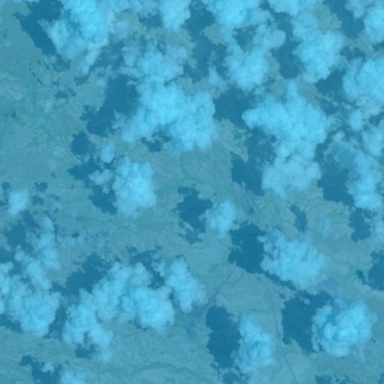}}
    & \viscell{{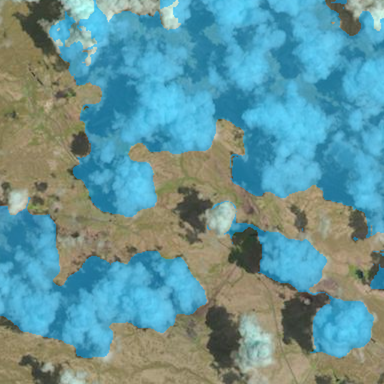}}
    & \viscell{{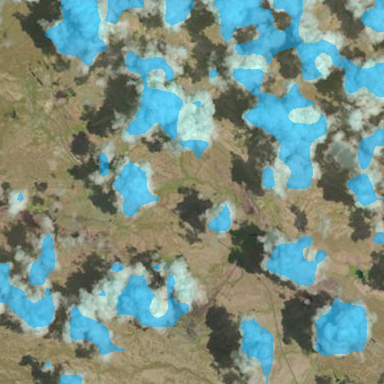}}%
}

\newcommand{\visrowhrcwhu}{%
    \viscell{{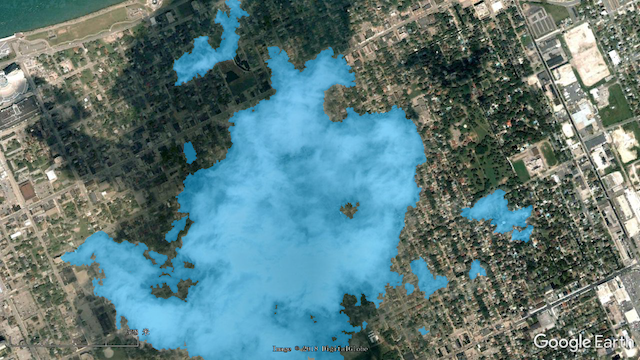}}
    & \viscell{{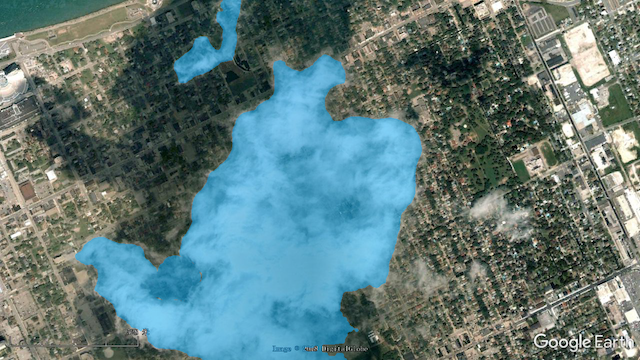}}
    & \viscell{{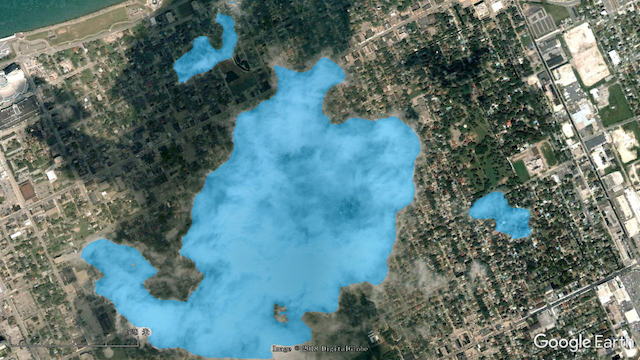}}
    & \viscell{{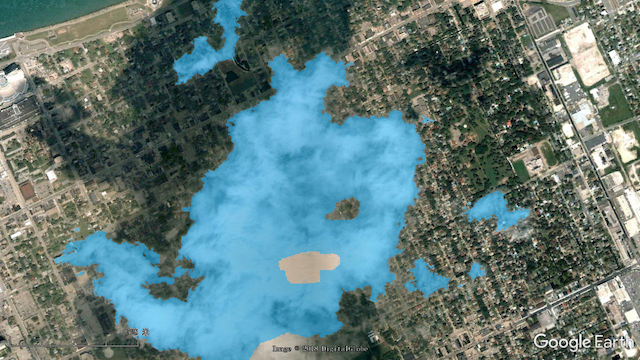}}
    & \viscell{{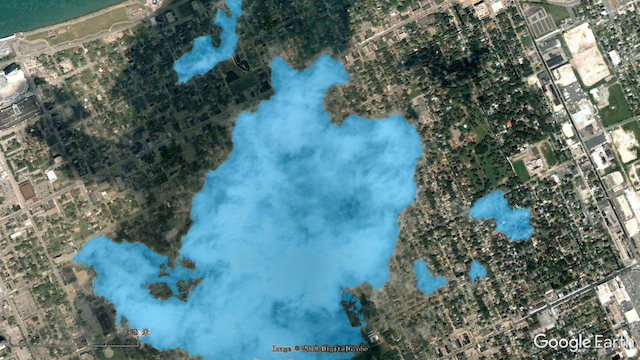}}
    & \viscell{{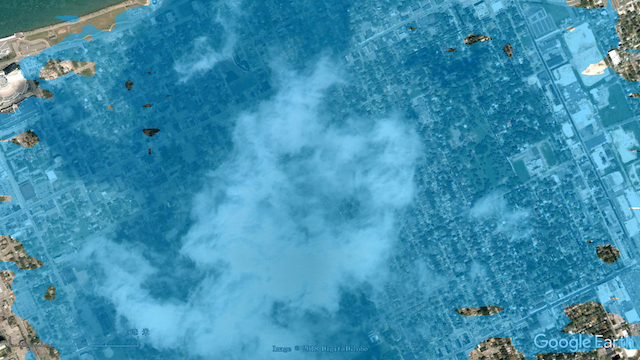}}
    & \viscell{{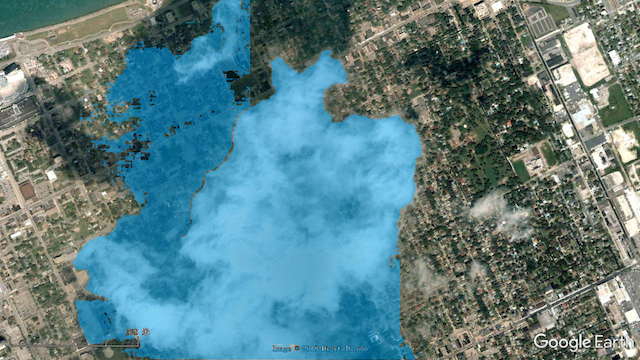}}
    & \viscell{{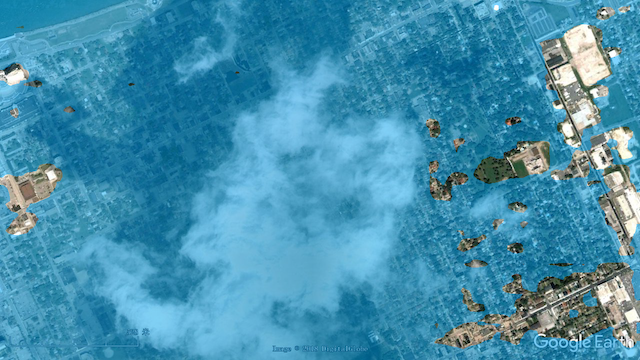}}
    & \viscell{{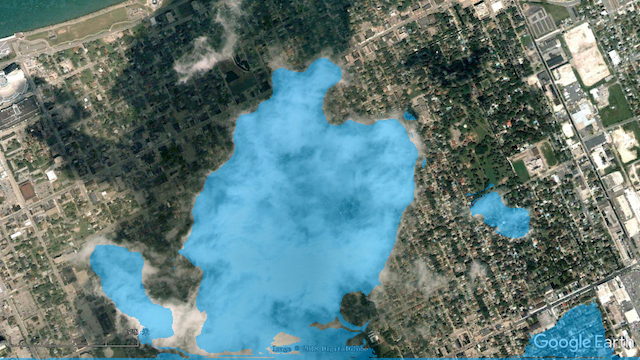}}
    & \viscell{{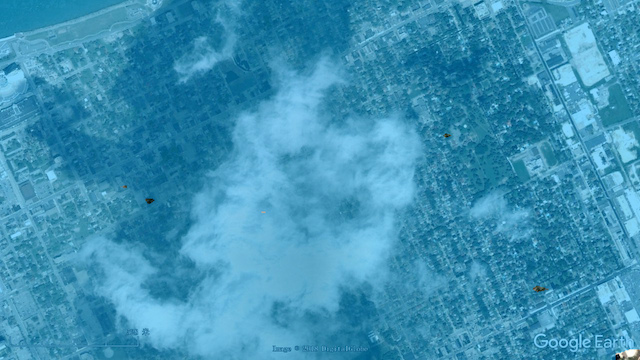}}%
}

\newcommand{\visrowsparcs}{%
    \viscell{{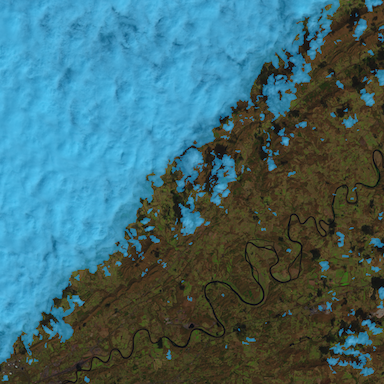}}
    & \viscell{{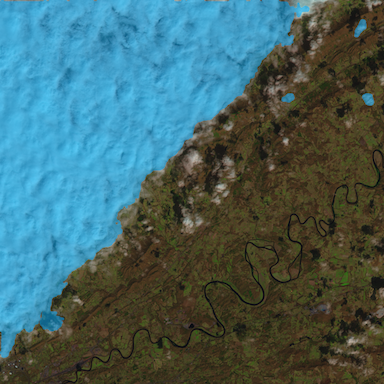}}
    & \viscell{{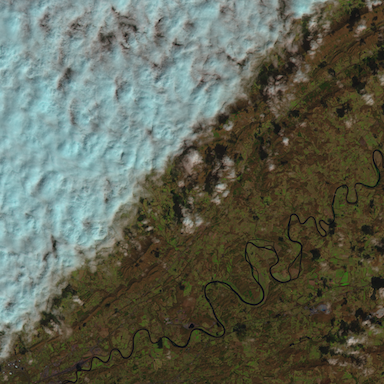}}
    & \viscell{{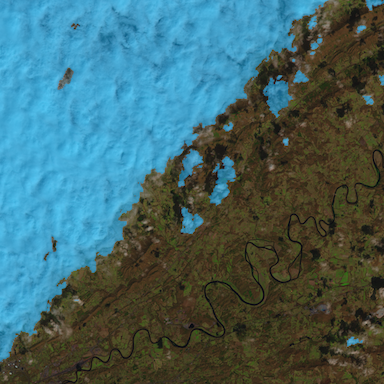}}
    & \viscell{{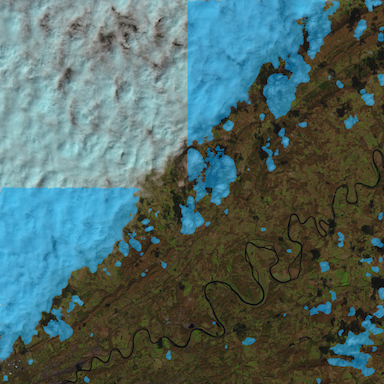}}
    & \viscell{{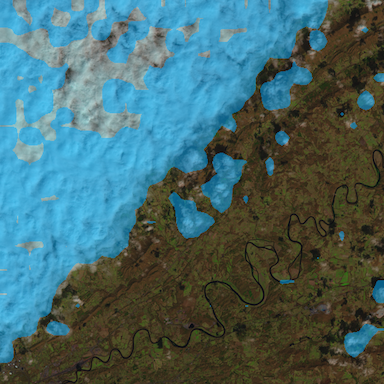}}
    & \viscell{{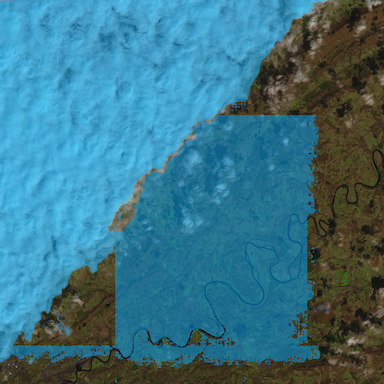}}
    & \viscell{{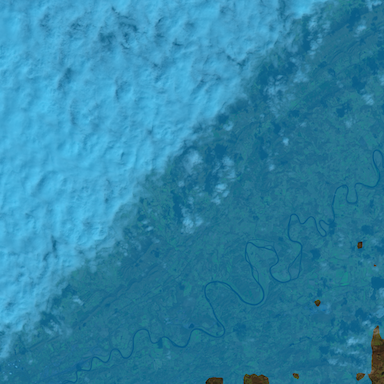}}
    & \viscell{{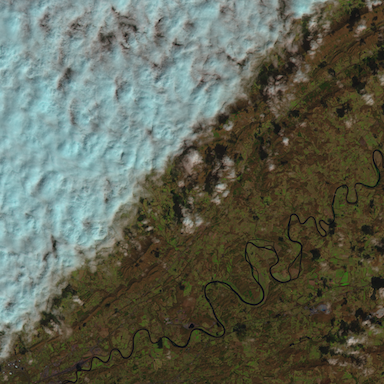}}
    & \viscell{{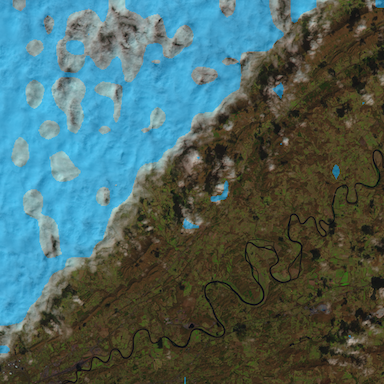}}%
}

% Number of visualization columns
\newlength{\viscellwidth}

% Calculate available width per cell
\setlength{\viscellwidth}{%
    \dimexpr(\textwidth - 18\tabcolsep)/10\relax
}

% Visualization cell
\newcommand{\viscell}[1]{%
    \includegraphics[
        width=\viscellwidth
    ]{#1}%
}

% Reusable grid
\newcommand{\comparisonGrid}{%
    \begingroup
    \centering
    \setlength{\tabcolsep}{2pt}
    \renewcommand{\arraystretch}{1.05}
    \scriptsize
    
\resizebox{\textwidth}{!}{%
    \begin{tabular}{@{}*{10}{c}@{}}

        \tiny\textbf{Image + GT}
        & \tiny\textbf{CLIP+SAM2}
        & \tiny\textbf{SAM3}
        & \tiny\textbf{SegEarth-OV}
        & \tiny\textbf{SegEarth-OV3}
        & \tiny\textbf{DINOv2+text}
        & \tiny\textbf{LangRS}
        & \tiny\textbf{GSNet}
        & \tiny\textbf{OpenSeg} 
        & \tiny\textbf{GEM} \\

        \visrowrice \\
        \visrowcloud \\
        \visrowsparcs \\
        \visrowhrcwhu \\

    \end{tabular}%
    }%
    \endgroup
}
\begin{figure*}[]
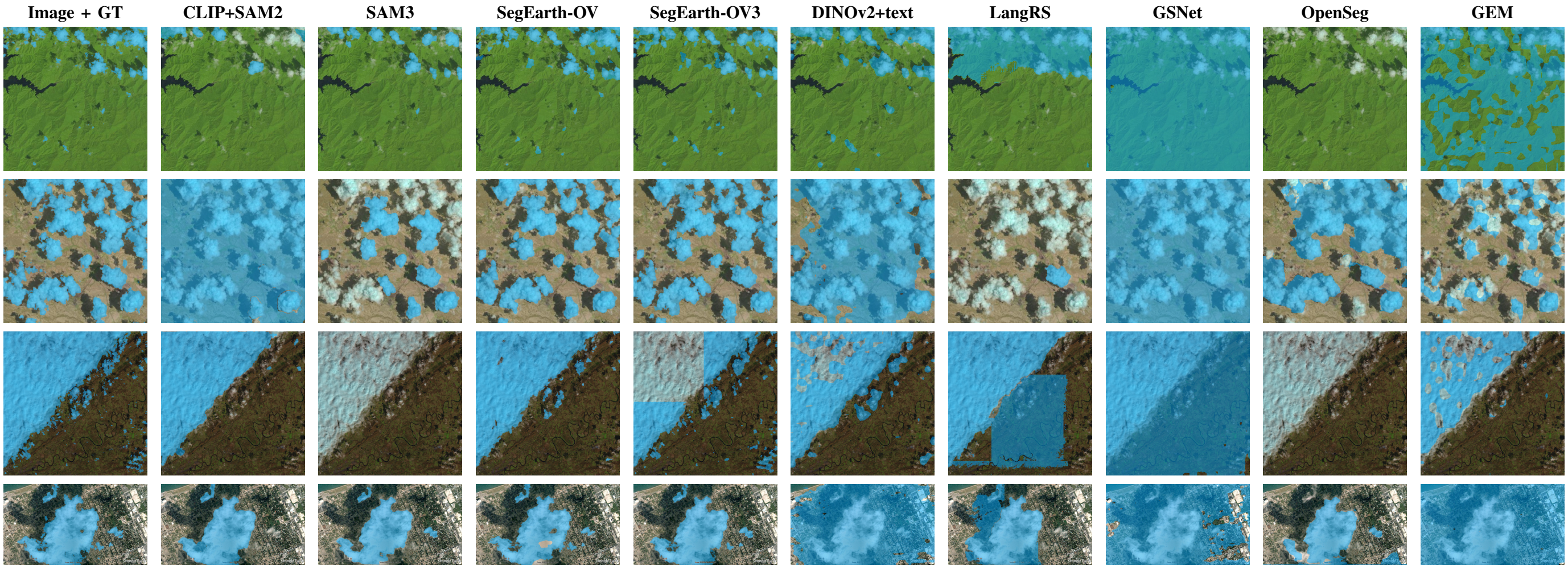

    \centering

    \comparisonGrid

    \caption{Qualitative comparison of zero-shot OVSS methods for binary cloud detection across four datasets: RICE, 95-Cloud, SPARCS, and HRC\_WHU with one representative example per dataset (row). The first column shows the original images overlaid with their corresponding ground-truth annotations, while the remaining columns show the cloud masks predicted by the evaluated OVSS methods. Cloud regions are highlighted in \textcolor{cloudblue}{blue}, while background regions are left uncoloured for clearer visualisation.}

    \label{fig:qualitative_comparison_1}
\end{figure*}

% Image cell definition (add to preamble if not already defined)
\newcommand{\viscellmulti}[1]{%
    \includegraphics[
        width=0.135\textwidth,
        height=0.135\textwidth,
        keepaspectratio
    ]{#1}%
}

\begin{figure*}[]
    \centering
    \setlength{\tabcolsep}{1pt}
    \renewcommand{\arraystretch}{1.05}

    \resizebox{\textwidth}{!}{%
    \begin{tabular}{@{}ccccc|cc@{}}
        \scriptsize \textbf{Image + GT}
        & \scriptsize \textbf{CLIP+SAM2}
        & \scriptsize \textbf{SegEarth-OV}
        & \scriptsize \textbf{SegEarth-OV3}
        & \scriptsize \textbf{SAM3}
        & \scriptsize \textbf{ChatGPT} 
        & \scriptsize \textbf{Gemini}\\
        
        % Example 1
        \viscellmulti{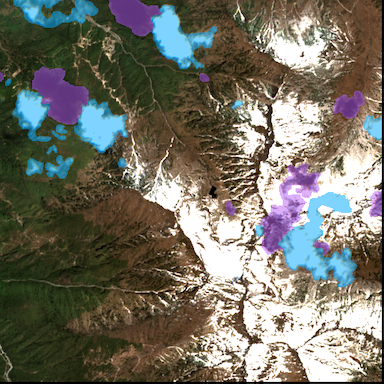}
        & \viscellmulti{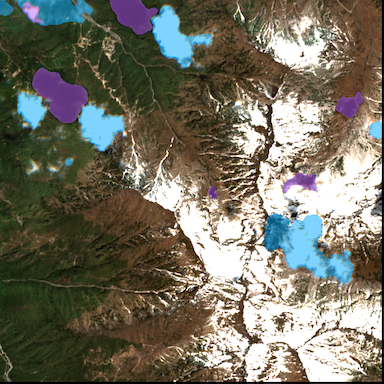}
        & \viscellmulti{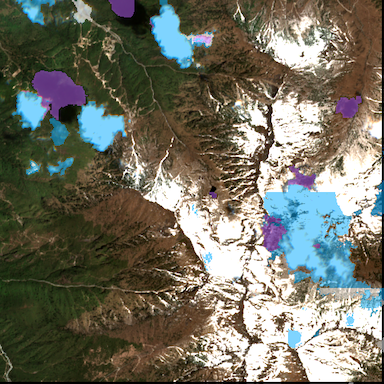}
        & \viscellmulti{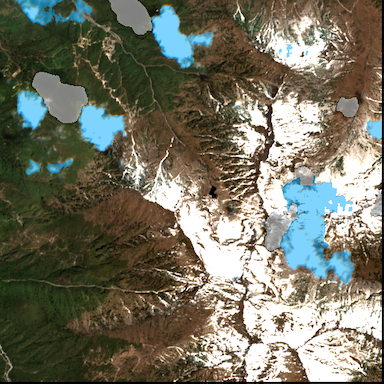}
        & \viscellmulti{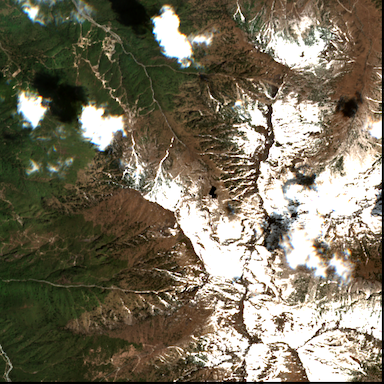}
        & \viscellmulti{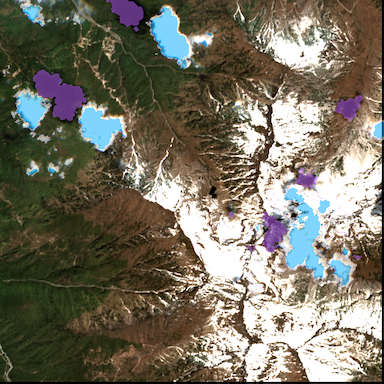}
        & \viscellmulti{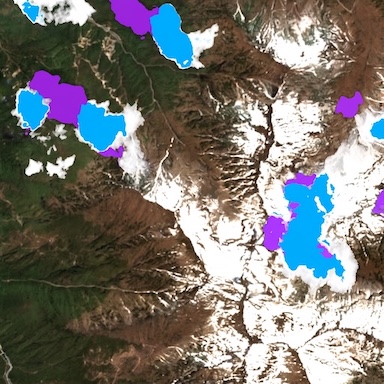} \\

        % Example 2
        \viscellmulti{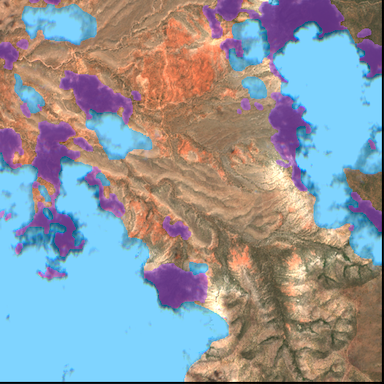}
        & \viscellmulti{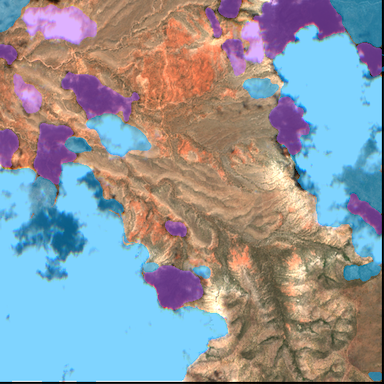}
        & \viscellmulti{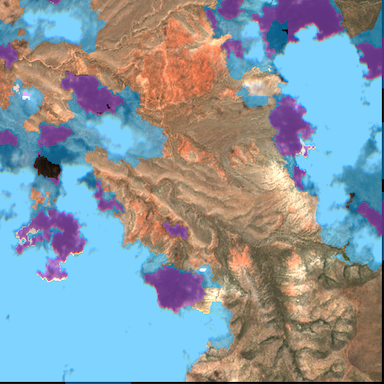}
        & \viscellmulti{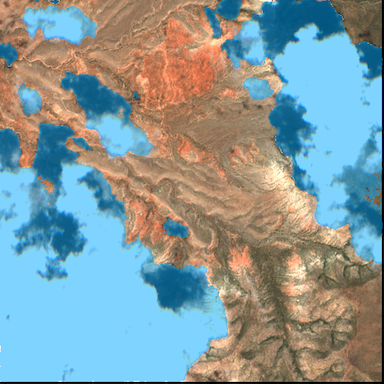}
        
        & \viscellmulti{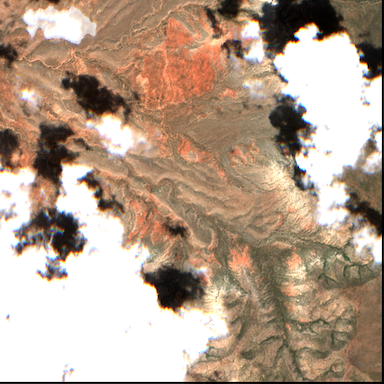}
        
        & \viscellmulti{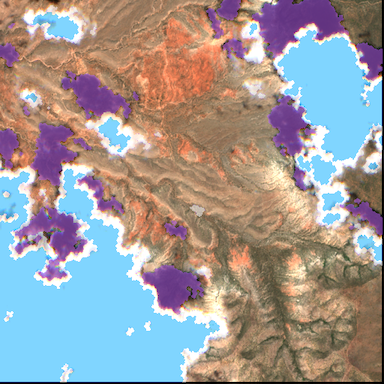} 
        & \viscellmulti{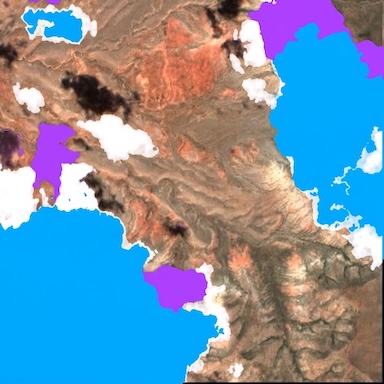}\\

        % Example 4
        %\viscellmulti{Figures/PNG/multiclass_vis/640_ROI_00974__20190821T053641_20190821T053749_T44UNE.png}
        \viscellmulti{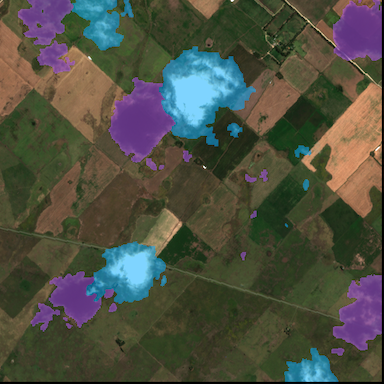}
        & \viscellmulti{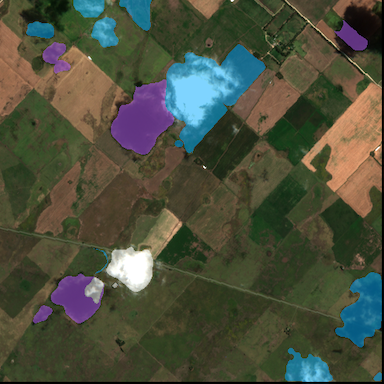}
        & \viscellmulti{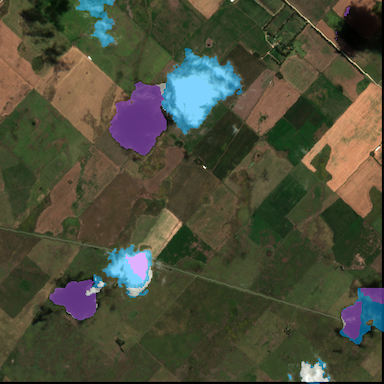}
        & \viscellmulti{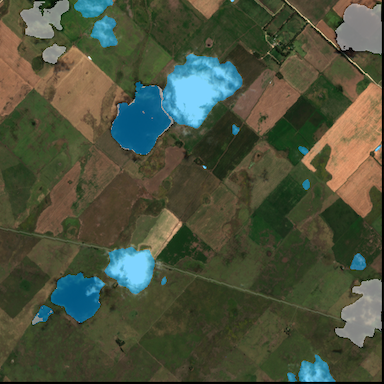}
        
        & \viscellmulti{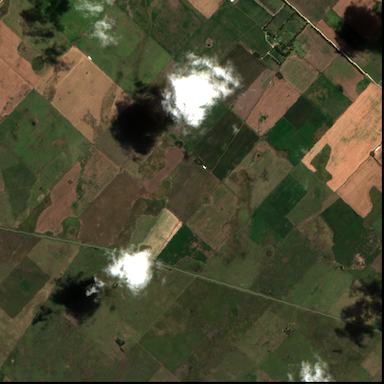}
        
        & \viscellmulti{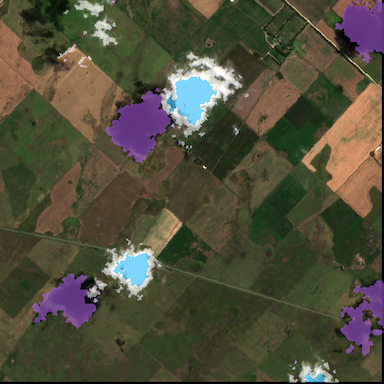} 
        & \viscellmulti{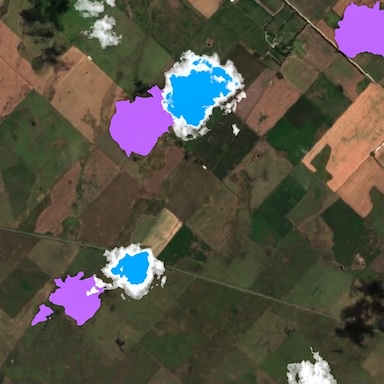} \\

        % Example 6
        \viscellmulti{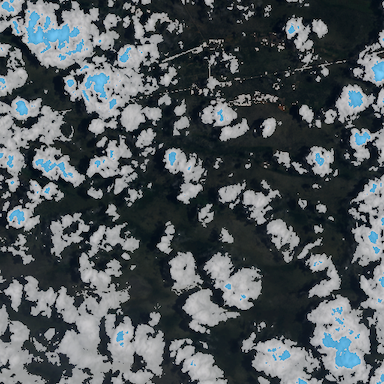}
        & \viscellmulti{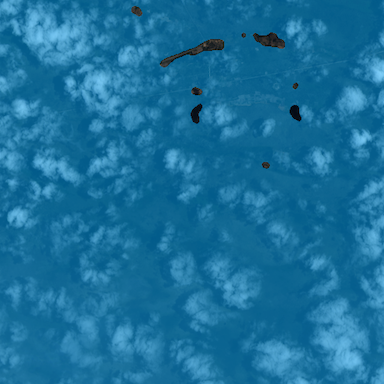}
        & \viscellmulti{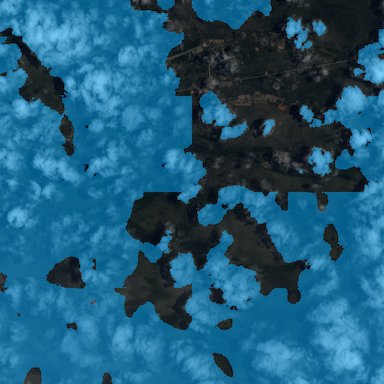}
        & \viscellmulti{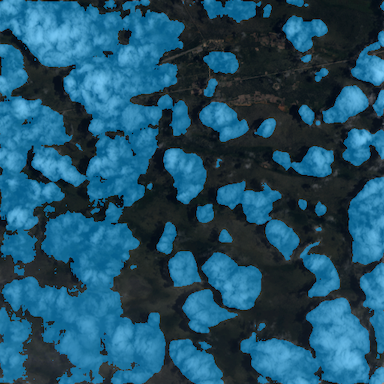}
        
       & \viscellmulti{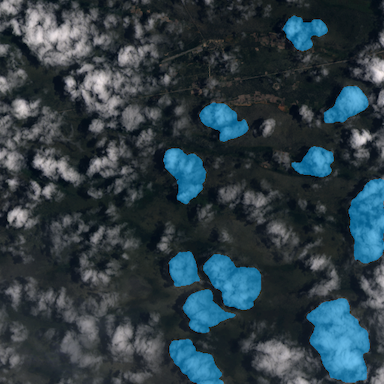}
        
        & \viscellmulti{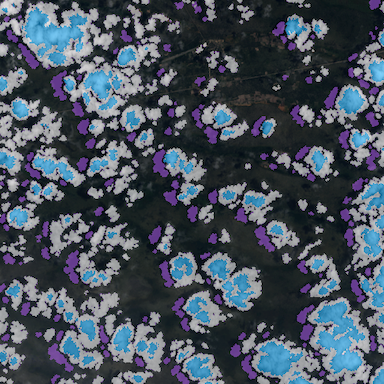} 
        & \viscellmulti{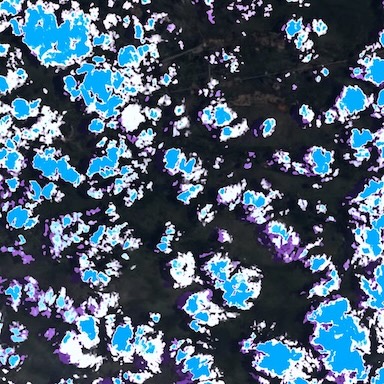} \\

    \end{tabular}
    }
    \caption{\textbf{Qualitative comparison of multiclass cloud segmentation across four open-source OVSS models and two closed-source models.} The first three rows show examples from the CloudSEN12 dataset, while the fourth row presents an example from the FG-OVSSRS dataset. Columns display the original RGB images with ground-truth (GT) annotations, and segmentation predictions from four open-source OVSS models (CLIP+SAM2, SegEarth-OV, SegEarth-OV3, and SAM3) and two closed-source models (ChatGPT and Gemini). Ground-truth and predicted segmentation masks are overlaid on the original RGB images. The semantic classes are thick clouds (\textcolor{cloudblue}{blue}), thin clouds (\textcolor{cloudgrey}{grey}), and cloud shadows (\textcolor{shadowpurple}{purple}). Note that the open-source OVSS methods produce segmentation predictions, whereas ChatGPT and Gemini generate segmentation-style overlays from image-based instructions.}
    \label{fig:multiclass_qualitative_comparison}
\end{figure*}

The exact textual prompts used for binary and multiclass segmentation are detailed in \autoref{tab:prompt_configurations}. In both settings, the background class remains unchanged. All models are evaluated using the same input images and inference settings as in the main experiments, with only the textual prompts varied across configurations. Model parameters remain frozen, and no task-specific adaptation or prompt optimisation is performed on the evaluation datasets.

\autoref{tab:prompt_sensitivity} presents the performance of the four evaluated models under the different prompt configurations. The results demonstrate differences in prompt sensitivity across models and segmentation settings. In particular, SAM3 benefits from \textit{Visual Proxy} prompts, achieving improvements of 2.97 and 7.14 points over the baseline in binary and multiclass segmentation, respectively. Unlike semantic descriptions, these prompts refer to general visual characteristics rather than explicitly identifying cloud-related categories. Their effectiveness therefore reflects the response of SAM3 to appearance-oriented concepts. Similarly, SegEarth-OV3, which builds upon SAM3, shows improved performance with alternative prompt formulations. Specifically, for the multiclass setting, where \textit{Descriptive} and \textit{Visual Proxy} prompts improve performance by 4.66 and 4.11 points, respectively.

In contrast, CLIP+SAM2 and SegEarth-OV exhibit different prompt sensitivity patterns. Both approaches rely on CLIP-based text representations and perform better when prompts include semantic concepts associated with the target classes, such as \textit{``clouds''}. In particular, \textit{Visual Proxy} prompts lead to a significant performance degradation in binary segmentation and multiclass segmentation for both models. Nevertheless, when visual attributes are paired with the semantic target noun, SegEarth-OV improves its binary segmentation performance by 3.46 and 1.77 points under \textit{Appearance} and \textit{Descriptive} prompts, respectively. For multiclass segmentation, \textit{Descriptive} prompts improve performance by 2.36 and 3.81 points for CLIP+SAM2 and SegEarth-OV, respectively.

\begin{table}[]
\centering
\caption{\textbf{Quantitative prompt sensitivity analysis for binary and multiclass open-vocabulary cloud segmentation.} Results are averaged across all evaluated datasets (gIoU): Binary reports mean cloud-class gIoU (\%) across six datasets and Multiclass reports mean 4-class gIoU (\%) across CloudSEN12 and FG-OVSSRS datasets. Differences relative to the baseline prompt configuration are shown in parentheses, with negative and positive changes highlighted in red and green, respectively. Values are rounded to two decimal places; differences are computed before rounding.}
\label{tab:prompt_sensitivity}

\setlength{\tabcolsep}{4pt}
\renewcommand{\arraystretch}{1.2}
\footnotesize

\resizebox{\columnwidth}{!}{%
\begin{tabular}{@{}llll@{}}
\toprule
\textbf{Model}
& \textbf{Prompt}
& \textbf{Binary}
& \textbf{Multiclass} \\
\midrule

\multirow{5}{*}{SAM3~\cite{carion2026sam}}
& Baseline
& 34.16 & 17.19 \\
& Semantic
& 15.66 {\scriptsize\textcolor{lightred}{(-18.51)}}
& 16.21 {\scriptsize\textcolor{lightred}{(-0.98)}} \\
& Appearance
& 30.35 {\scriptsize\textcolor{lightred}{(-3.82)}}
& 11.67 {\scriptsize\textcolor{lightred}{(-5.53)}} \\
& Descriptive
& 33.37 {\scriptsize\textcolor{lightred}{(-0.79)}}
& 18.08 {\scriptsize\textcolor{lightgreen}{(+0.89)}} \\
& Visual Proxy
& \textbf{37.14} {\scriptsize\textcolor{lightgreen}{(+2.97)}}
& \textbf{24.34} {\scriptsize\textcolor{lightgreen}{(+7.14)}} \\

\midrule

\multirow{5}{*}{SegEarth-OV3~\cite{li2025segearthov3}}
& Baseline
& \textbf{55.55} & 19.79 \\
& Semantic
& 40.24 {\scriptsize\textcolor{lightred}{(-15.31)}}
& 17.77 {\scriptsize\textcolor{lightred}{(-2.02)}} \\
& Appearance
& 53.08 {\scriptsize\textcolor{lightred}{(-2.47)}}
& 15.76 {\scriptsize\textcolor{lightred}{(-4.03)}} \\
& Descriptive
& 53.06 {\scriptsize\textcolor{lightred}{(-2.49)}}
& \textbf{24.46} {\scriptsize\textcolor{lightgreen}{(+4.66)}} \\
& Visual Proxy
& 50.01 {\scriptsize\textcolor{lightred}{(-5.54)}}
& 23.91 {\scriptsize\textcolor{lightgreen}{(+4.11)}} \\

\midrule

\multirow{5}{*}{CLIP+SAM2~\cite{sosa2026enabling}}
& Baseline
& \textbf{46.34} & 21.44 \\
& Semantic
& 42.11 {\scriptsize\textcolor{lightred}{(-4.23)}}
& 20.90 {\scriptsize\textcolor{lightred}{(-0.55)}} \\
& Appearance
& 42.97 {\scriptsize\textcolor{lightred}{(-3.37)}}
& 16.37 {\scriptsize\textcolor{lightred}{(-5.08)}} \\
& Descriptive
& 45.76 {\scriptsize\textcolor{lightred}{(-0.58)}}
& \textbf{23.80} {\scriptsize\textcolor{lightgreen}{(+2.36)}} \\
& Visual Proxy
& 8.75 {\scriptsize\textcolor{lightred}{(-37.59)}}
& 11.20 {\scriptsize\textcolor{lightred}{(-10.25)}} \\

\midrule

\multirow{5}{*}{SegEarth-OV~\cite{li2025segearth}}
& Baseline
& 51.75 & 23.78 \\
& Semantic
& 48.81 {\scriptsize\textcolor{lightred}{(-2.94)}}
& 22.64 {\scriptsize\textcolor{lightred}{(-1.14)}} \\
& Appearance
& \textbf{55.21} {\scriptsize\textcolor{lightgreen}{(+3.46)}}
& 17.84 {\scriptsize\textcolor{lightred}{(-5.94)}} \\
& Descriptive
& 53.52 {\scriptsize\textcolor{lightgreen}{(+1.77)}}
& \textbf{27.59} {\scriptsize\textcolor{lightgreen}{(+3.81)}} \\
& Visual Proxy
& 16.35 {\scriptsize\textcolor{lightred}{(-35.40)}}
& 12.69 {\scriptsize\textcolor{lightred}{(-11.09)}} \\

\bottomrule
\end{tabular}%
}
\end{table}

\noindent\textbf{Closed-source models.} As an exploratory extension of our multiclass evaluation, we also present qualitative predictions generated using closed-source models through the ChatGPT and Gemini interfaces. Specifically, we used ChatGPT with Pro reasoning capabilities and Gemini 3.6 Flash. Unlike the open-source OVSS approaches evaluated quantitatively, these models generate segmentation overlays directly from image-based instructions. Thus, this comparison is intended solely to illustrate their potential for fine-grained cloud segmentation rather than establish a direct performance ranking. 

As these predictions were generated through user interfaces rather than direct API calls, the underlying image-generation model versions could not be independently verified. We therefore limit the analysis to qualitative examples, comparing the generated overlays with those obtained from open-source OVSS approaches, as shown in the last two columns of~\autoref{fig:multiclass_qualitative_comparison}. To generate these predictions, we provide the corresponding input image along with the following text instruction:

\begin{tcolorbox}[
    colback=gray!8,
    colframe=gray!35,
    boxrule=0.5pt,
    arc=2mm,
    left=8pt,
    right=8pt,
    top=6pt,
    bottom=6pt
]
\small
\textit{Produce a segmentation map for this image to segment thick clouds, thin clouds, and cloud shadows. Use the following colour palette for the classes: cloud shadow (190, 100, 255), thin cloud (255, 255, 255), and thick cloud (0, 170, 255). After generating the predictions, overlay the segmentation mask on the input image using an alpha value of 0.5.}
\end{tcolorbox}

Although the predictions generated by these models appear visually closer to the ground truth in some examples than those produced by the evaluated OVSS approaches, they still exhibit difficulties in distinguishing fine-grained cloud categories. Confusion between thick and thin clouds remains evident, as illustrated by the example in the last row of \autoref{fig:multiclass_qualitative_comparison}. Moreover, their closed-source nature limits transparency regarding model architectures and inference procedures. At the same time, computational costs may increase with more demanding reasoning configurations. Nevertheless, a full quantitative evaluation would be necessary to determine whether the observed qualitative differences translate into improved segmentation performance.

\section{Conclusions}

In this work, we evaluated the zero-shot cloud segmentation capabilities of twelve OVSS approaches across six diverse EO datasets. Our results demonstrate that current OVSS models can identify cloud regions without task-specific training, although their performance varies across models and datasets. While EO-adapted approaches achieve competitive results, their relative performance depends on the evaluated datasets and may also be influenced by architectural differences.

Despite promising results for binary cloud detection, our multiclass evaluation shows limitations in distinguishing between thick clouds, thin clouds, and cloud shadows. In particular, all evaluated approaches struggle to identify thin clouds and cloud shadows, highlighting the gap between coarse cloud detection and fine-grained cloud characterisation. Our controlled prompt sensitivity analysis also highlights model-dependent responses to alternative textual descriptions, with appearance-oriented prompts benefiting certain approaches while degrading others.

Overall, our study demonstrates both the potential and current limitations of OVSS for cloud detection with EO imagery. The results highlight the need for improved fine-grained semantic discrimination, greater robustness to prompt formulation, and more consistent evaluation protocols that account for annotation uncertainty. By providing a complete benchmark across diverse models, datasets, and prompt configurations, we aim to support future research towards more reliable and generalisable zero-shot OVSS for cloud detection.
 
\vspace{1em}

\noindent\textbf{Acknowledgments.} We thank Ons AOUEDI, Hieu TRAN DINH, Ti Ti NGUYEN, Muhammad AWAIS, and Vu Nguyen HA from the Signal Processing and Communications group at the University of Luxembourg for their valuable discussions. We also thank Tereza Pohankova from Hydrosat for her feedback on the visual materials presented in this paper. This work was supported by the Luxembourg Ministry of the Economy (MECO). All experiments were conducted on the MeluXina supercomputer. We extend our special thanks to the LuxProvide team for their technical support.

\small
\bibliographystyle{IEEEtranN}
\bibliography{references}

\end{document}